# Learning with Augmented Features for Heterogeneous Domain Adaptation

**Lixin Duan**  S080003@ntu.edu.sg
**Dong Xu**  DongXu@ntu.edu.sg
**Ivor W. Tsang**  IvorTsang@ntu.edu.sg
School of Computer Engineering, Nanyang Technological University, 50 Nanyang Avenue, Singapore 639798

## Abstract

We propose a new learning method for heterogeneous domain adaptation (HDA), in which the data from the source domain and the target domain are represented by heterogeneous features with different dimensions. Using two different projection matrices, we first transform the data from two domains into a common subspace in order to measure the similarity between the data from two domains. We then propose two new feature mapping functions to augment the transformed data with their original features and zeros. The existing learning methods (e.g., SVM and SVR) can be readily incorporated with our newly proposed augmented feature representations to effectively utilize the data from both domains for HDA. Using the hinge loss function in SVM as an example, we introduce the detailed objective function in our method called Heterogeneous Feature Augmentation (HFA) for a linear case and also describe its kernelization in order to efficiently cope with the data with very high dimensions. Moreover, we also develop an alternating optimization algorithm to effectively solve the nontrivial optimization problem in our HFA method. Comprehensive experiments on two benchmark datasets clearly demonstrate that HFA outperforms the existing HDA methods.

## 1. Introduction

In real-world applications, it is often expensive and time-consuming to collect the labeled data. Transfer learning (a.k.a., domain adaptation), as a new machine learning strategy, has attracted growing attention because it can learn robust classifiers with very few labeled data from the target domain by leveraging a large amount of labeled data from other existing domains (a.k.a., source domains).

Domain adaptation methods have been successfully used for different research fields such as natural language processing and computer vision (Blitzer et al., 2006; 2007; Daumé III, 2007; Duan et al., 2010; 2012b;a; Wu & Dietterich, 2004). However, all those methods assume that the data from different domains are represented by the same type of features with the same dimension. Thus, they cannot deal with the problem where the dimensions of data from the source and target domains are different, which is known as heterogeneous domain adaptation (HDA) (Dai et al., 2009; Yang et al., 2009).

In the literature, a few works have been proposed for the HDA problem. Dai et al. (2009) proposed to learn a feature translator between the source and target domains by assuming that the data from both domains share co-occurrence attributes (i.e., text data). The same assumption was also used in (Yang et al., 2009; Zhu et al., 2011) for text-aid image clustering and classification. However, this assumption may not be well satisfied in many applications such as the object recognition task where only visual features are used. Based on structural correspondence learning (Blitzer et al., 2006), two methods (Prettenhofer & Stein, 2010; Wei & Pal, 2010) were recently proposed to extract the so-called *pivot* features from the source and target domains, which is specifically designed for the cross-language text classification task. And these pivot features are constructed by text words which have explicit semantic meanings.

For more general HDA tasks, Shi et al. (2010) proposed a method called Heterogeneous Spectral Mapping

(HeMap) to discover a common feature subspace by learning two feature mapping matrices as well as the optimal projection of the data from both domains, in which the valuable label information is not exploited. Harel and Mannor (2011) learned rotation matrices to match source data distributions to that of the target domain. However, this method does not use the valuable training labels, either. Wang et al. (2011) used the class labels of the training data to learn the manifold alignment by simultaneously maximizing the intra-domain similarity and the inter-domain dissimilarity. By kernelizing the method in (Saenko et al., 2010), Kulis et al. (2011) proposed to learn an asymmetric kernel transformation to transfer feature knowledge between the data from the source and target domains.

In this work, we propose a new method called Heterogeneous Feature Augmentation (HFA) for heterogeneous domain adaptation. Considering the data from different domains are represented by features with different dimensions, we first transform the data from the source and target domains into a common subspace by using two different projection matrices $\mathbf{P}$ and $\mathbf{Q}$. Then, we propose two new feature mapping functions to augment the transformed data with their original features and zeros. With the new augmented feature representations, we propose to learn the projection matrices $\mathbf{P}$ and $\mathbf{Q}$ by using the standard SVM with the hinge loss function in a linear case. We also describe its kernelization in order to efficiently cope with the data with very high dimension. To simplify the nontrivial optimization problem in HFA, we introduce an intermediate variable $\mathbf{H}$ called as a transformation metric to combine $\mathbf{P}$ and $\mathbf{Q}$. We then develop an alternating optimization algorithm to simultaneously solve for the dual problem of SVM and the optimal transformation metric $\mathbf{H}$.

We summarize the main contributions of this work:

- The newly proposed augmented features in our HFA method can be readily incorporated into different methods (e.g., SVM and SVR) to effectively utilize the patterns from two domains, making them applicable to the HDA task.

- We simplify the nontrivial optimization problem by defining a transformation metric $\mathbf{H}$ and develop an effective alternating optimization algorithm. With the introduction of $\mathbf{H}$, we do not explicitly solve for $\mathbf{P}$ and $\mathbf{Q}$, which makes the common subspace *invisible* to us.

- Promising results on two benchmark datasets clearly demonstrate the effectiveness of HFA for object recognition and text categorization.

## 2. Kernel Learning for Heterogeneous Domain Adaptation

In the remainder of this paper, we use the superscript $'$ to denote the transpose of a vector or a matrix. We define $\mathbf{I}_n$ as the $n \times n$ identity matrix and $\mathbf{O}_{n \times m}$ as a $n \times m$ matrix of all zeros. We also define $\mathbf{0}_n, \mathbf{1}_n \in \mathbb{R}^n$ as the $n \times 1$ column vectors of all zeros and all ones, respectively. The inequality $\mathbf{a} \leq \mathbf{b}$ means that $a_i \leq b_i$ for $i = 1, \ldots, n$. Moreover, $\mathbf{a} \circ \mathbf{b}$ denotes the element-wise product between vectors $\mathbf{a}$ and $\mathbf{b}$, i.e., $\mathbf{a} \circ \mathbf{b} = [a_1 b_1, \ldots, a_n b_n]'$. And $\mathbf{H} \succeq 0$ means that the matrix $\mathbf{H}$ is positive semidefinite.

In this work, we assume there are only one source domain and one target domain. For some given class, we are provided with a set of labeled training samples $\{(\mathbf{x}_i^s, y_i^s)|_{i=1}^{n_s}\}$ from the source domain as well as a limited number of labeled samples $\{(\mathbf{x}_i^t, y_i^t)|_{i=1}^{n_t}\}$ from the target domain, where $y_i^s$ and $y_i^t$ are the labels of the samples $\mathbf{x}_i^s$ and $\mathbf{x}_i^t$, respectively, and $y_i^s, y_i^t \in \{1, -1\}$. The dimensions of $\mathbf{x}_i^s$ and $\mathbf{x}_i^t$ are $d_s$ and $d_t$, respectively. Note that in the HDA problem, $d_s \neq d_t$.

### 2.1. Heterogeneous Feature Augmentation

Daume III (2007) proposed Feature Replication (FR) to augment the original feature space $\mathbb{R}^d$ into a larger space $\mathbb{R}^{3d}$ by replicating the source and target data for homogeneous domain adaptation. Specifically, for any data point $\mathbf{x} \in \mathbb{R}^d$, the feature mapping functions $\varphi_s$ and $\varphi_t$ for the source and target domains are defined as $\varphi_s(\mathbf{x}) = [\mathbf{x}', \mathbf{x}', \mathbf{0}_d']'$ and $\varphi_t(\mathbf{x}) = [\mathbf{x}', \mathbf{0}_d, \mathbf{x}']'$. Note that it is not meaningful to directly use the method in (Daumé III, 2007) for the HDA task by simply padding zeros to make the dimensions of the data from two domains become the same, because there would be no correspondences between the heterogeneous features in this case.

To effectively utilize the heterogeneous features from two domains, we first introduce a common subspace for the source and target data for our HDA task, in which the heterogeneous features from two domains can be compared. We define the common subspace as $\mathbb{R}^{d_c}$, where any source sample $\mathbf{x}^s$ and target sample $\mathbf{x}^t$ can be projected onto it by using two projection matrices $\mathbf{P} \in \mathbb{R}^{d_c \times d_s}$ and $\mathbf{Q} \in \mathbb{R}^{d_c \times d_t}$, respectively. Note that promising results have been shown by incorporating original features into feature augmentation (Daumé III, 2007; Pan et al., 2010) to enhance the similarities between data from the same domain. Motivated by (Daumé III, 2007; Pan et al., 2010), we also incorporate original features in this work and then augment any source and target domain samples $\mathbf{x}^s \in \mathbb{R}^{d_s}$ and $\mathbf{x}^t \in \mathbb{R}^{d_t}$ by using our newly proposed feature mapping functions $\varphi_s$ and $\varphi_t$ as follows:

$$\varphi_s(\mathbf{x}^s) = \begin{bmatrix} \mathbf{P}\mathbf{x}^s \\ \mathbf{x}^s \\ \mathbf{0}_{d_t} \end{bmatrix} \text{ and } \varphi_t(\mathbf{x}^t) = \begin{bmatrix} \mathbf{Q}\mathbf{x}^t \\ \mathbf{0}_{d_s} \\ \mathbf{x}^t \end{bmatrix}. \quad (1)$$

After introducing $\mathbf{P}$ and $\mathbf{Q}$, the data from two domains can be readily compared in the common subspace. It is worth mentioning that our newly proposed augmented features for the source and target samples in (1) can be readily incorporated into different methods (e.g., SVM and SVR), making these methods applicable for the HDA problem.

In the next subsection, we take SVM with the hinge loss as a showcase of our **H**eterogeneous **F**eature **A**ugmentation method (HFA for short). As it is nontrivial to solve for the projection matrices $\mathbf{P}$ and $\mathbf{Q}$ in our learning problem, we simplify the optimization problem by introducing an intermediate variable $\mathbf{H} = [\mathbf{P}, \mathbf{Q}]'[\mathbf{P}, \mathbf{Q}]$ such that we only need to solve for $\mathbf{H}$ rather than $\mathbf{P}$ and $\mathbf{Q}$. In this way, the common subspace becomes *invisible* to us, which is therefore referred to as *latent common subspace* in this work.

### 2.2. Proposed Method

We define feature weight vector $\mathbf{w} = [\mathbf{w}_c', \mathbf{w}_s', \mathbf{w}_t']'$ for the augmented feature space, where $\mathbf{w}_c, \mathbf{w}_s$ and $\mathbf{w}_t$ are also weight vectors that are defined for the common subspace, the source domain and the target domain, respectively. We then propose to learn the projection matrices $\mathbf{P}$ and $\mathbf{Q}$ as well as the weight vector $\mathbf{w}$ by minimizing the structural risk functional of SVM. Formally, we present the formulation of our HFA method for the HDA problem as follows:

$$\min_{\mathbf{P},\mathbf{Q}} \min_{\mathbf{w}, b, \xi_i^s, \xi_i^t} \frac{1}{2}\|\mathbf{w}\|^2 + C\left(\sum_{i=1}^{n_s}\xi_i^s + \sum_{i=1}^{n_t}\xi_i^t\right), \quad (2)$$

$$\text{s.t.} \quad y_i^s(\mathbf{w}'\varphi_s(\mathbf{x}_i^s) + b) \geq 1 - \xi_i^s, \xi_i^s \geq 0; \quad (3)$$
$$y_i^t(\mathbf{w}'\varphi_t(\mathbf{x}_i^t) + b) \geq 1 - \xi_i^t, \xi_i^t \geq 0; \quad (4)$$
$$\|\mathbf{P}\|_F^2 \leq \lambda_p, \|\mathbf{Q}\|_F^2 \leq \lambda_q,$$

where $C > 0$ is a regularization parameter that regulates the loss on the training samples, and $\lambda_p, \lambda_q > 0$ are predefined to control the complexities of $\mathbf{P}$ and $\mathbf{Q}$, respectively.

To solve (2), we first derive the dual form of the inner optimization problem in (2) with respect to $\mathbf{w}, b, \xi_i^s$ and $\xi_i^t$. Specifically, we introduce dual variables $\{\alpha_i^s|_{i=1}^{n_s}\}$ and $\{\alpha_i^t|_{i=1}^{n_t}\}$ for the constraints in (3) and (4), respectively. By setting the derivatives of the Lagrangian of (2) with respect to $\mathbf{w}, b, \xi_i^s$ and $\xi_i^t$ to zeros, we obtain the Karush-Kuhn-Tucker (KKT) conditions as: $\mathbf{w} = \sum_{i=1}^{n_s}\alpha_i^s y_i^s \varphi_s(\mathbf{x}_i^s) + \sum_{i=1}^{n_t}\alpha_i^t y_i^t \varphi_t(\mathbf{x}_i^t)$, $\sum_{i=1}^{n_s}\alpha_i^s y_i^s + \sum_{i=1}^{n_t}\alpha_i^t y_i^t = 0$ and $0 \leq \alpha_i^s, \alpha_i^t \leq C$.

With the KKT conditions, we arrive at the alternative optimization problem as follows:

$$\min_{\mathbf{P},\mathbf{Q}} \max_{\boldsymbol{\alpha}} \quad \mathbf{1}'_{n_s+n_t}\boldsymbol{\alpha} - \frac{1}{2}(\boldsymbol{\alpha} \circ \mathbf{y})'\mathbf{K}_{\mathbf{P},\mathbf{Q}}(\boldsymbol{\alpha} \circ \mathbf{y}), \quad (5)$$

$$\text{s.t.} \quad \mathbf{y}'\boldsymbol{\alpha} = 0, \mathbf{0}_{n_s+n_t} \leq \boldsymbol{\alpha} \leq C\mathbf{1}_{n_s+n_t},$$
$$\|\mathbf{P}\|_F^2 \leq \lambda_p, \|\mathbf{Q}\|_F^2 \leq \lambda_q,$$

where $\boldsymbol{\alpha} = [\alpha_1^s, \ldots, \alpha_{n_s}^s, \alpha_1^t, \ldots, \alpha_{n_t}^t]'$ is a vector of the dual variables, $\mathbf{y} = [y_1^s, \ldots, y_{n_s}^s, y_1^t, \ldots, y_{n_t}^t]'$ is a label vector, $\mathbf{K}_{\mathbf{P},\mathbf{Q}} = \begin{bmatrix} \mathbf{X}_s'(\mathbf{I}_{n_s} + \mathbf{P}'\mathbf{P})\mathbf{X}_s & \mathbf{X}_s'\mathbf{P}'\mathbf{Q}\mathbf{X}_t \\ \mathbf{X}_t'\mathbf{Q}'\mathbf{P}\mathbf{X}_s & \mathbf{X}_t'(\mathbf{I}_{n_t} + \mathbf{Q}'\mathbf{Q})\mathbf{X}_t \end{bmatrix}$ is the derived kernel matrix.

A straightforward solution to the optimization problem in (5) would be to iteratively update one of the variables $\boldsymbol{\alpha}, \mathbf{P}$ and $\mathbf{Q}$ by fixing the others. However, the dimension of the common subspace (i.e., $d_c$) must be given beforehand in this case, and it is nontrivial to determine the optimal $d_c$. Observing that in the kernel matrix $\mathbf{K}_{\mathbf{P},\mathbf{Q}}$ in (5), the projection matrices $\mathbf{P}$ and $\mathbf{Q}$ always appear in the forms of $\mathbf{P}'\mathbf{P}, \mathbf{P}'\mathbf{Q}, \mathbf{Q}'\mathbf{P}$ and $\mathbf{Q}'\mathbf{Q}$, we then replace these multiplications by defining an intermediate variable $\mathbf{H} = [\mathbf{P}, \mathbf{Q}]'[\mathbf{P}, \mathbf{Q}] \in \mathbb{R}^{(d_s+d_t)\times(d_s+d_t)}$. Obviously, $\mathbf{H}$ is positive semidefinite, i.e., $\mathbf{H} \succeq 0$. With the introduction of $\mathbf{H}$, we can throw away the parameter $d_c$. Moreover, the common subspace becomes *latent*, because we do not need to explicitly solve for $\mathbf{P}$ and $\mathbf{Q}$ any more.

With the definition of $\mathbf{H}$, we convert the optimization problem in (5) to the final formulation of our proposed HFA method as follows:

$$\min_{\mathbf{H}\succeq 0} \max_{\boldsymbol{\alpha}} \quad \mathbf{1}'_{n_s+n_t}\boldsymbol{\alpha} - \frac{1}{2}(\boldsymbol{\alpha} \circ \mathbf{y})'\mathbf{K}_{\mathbf{H}}(\boldsymbol{\alpha} \circ \mathbf{y}), \quad (6)$$

$$\text{s.t.} \quad \mathbf{y}'\boldsymbol{\alpha} = 0, \mathbf{0}_{n_s+n_t} \leq \boldsymbol{\alpha} \leq C\mathbf{1}_{n_s+n_t}, \text{trace}(\mathbf{H}) \leq \lambda,$$

where $\mathbf{K}_{\mathbf{H}} = \begin{bmatrix} \mathbf{X}_s'\mathbf{X}_s + \mathbf{L}_s'\mathbf{H}\mathbf{L}_s & \mathbf{L}_s'\mathbf{H}\mathbf{L}_t \\ \mathbf{L}_t'\mathbf{H}\mathbf{L}_s & \mathbf{X}_t'\mathbf{X}_t + \mathbf{L}_t'\mathbf{H}\mathbf{L}_t \end{bmatrix}$, $\mathbf{L}_s = \begin{bmatrix} \mathbf{I}_{d_s} \\ \mathbf{0}_{d_t \times d_s} \end{bmatrix}\mathbf{X}_s$, $\mathbf{L}_t = \begin{bmatrix} \mathbf{0}_{d_s \times d_t} \\ \mathbf{I}_{d_t} \end{bmatrix}\mathbf{X}_t$ and $\lambda = \lambda_p + \lambda_q$. Note that given $\boldsymbol{\alpha}$, the optimization problem in (6) becomes the following Semidefinite Programming (SDP) problem (Vandenberghe & Boyd, 1996) by defining $\boldsymbol{\beta} = \boldsymbol{\alpha} \circ \mathbf{y}$:

$$\min_{\mathbf{H}\succeq 0} -\frac{1}{2}\boldsymbol{\beta}'\mathbf{K}_{\mathbf{H}}\boldsymbol{\beta}, \quad \text{s.t. trace}(\mathbf{H}) \leq \lambda. \quad (7)$$

Thus far, we have successfully converted our original HDA problem, which learns two projection matrices $\mathbf{P}$ and $\mathbf{Q}$, into a new problem of learning a *transformation metric* $\mathbf{H}$. We emphasize that this new problem has two main advantages: i) it avoids determining the optimal dimension of the common subspace beforehand; and ii) as the common subspace

Learning with Augmented Features for Heterogeneous Domain Adaptation

becomes latent after the introduction of $\mathbf{H}$, we only have to optimize $\boldsymbol{\alpha}$ and $\mathbf{H}$ for our proposed method.

**Discussion:** There are two major limitations to the current formulation of HFA in (6): i) The transformation metric $\mathbf{H}$ is linear, which may not be effective for some tasks. ii) The size of $\mathbf{H}$ grows with the dimensions of the source and target data (i.e., $d_s$ and $d_t$). Therefore, it is computationally infeasible to learn the linear metric $\mathbf{H}$ in the SDP problem (7) for some real-world applications (e.g., text categorization) with very high dimensional data. In order to effectively deal with high dimensional data, inspired by (Kulis et al., 2011), in the next subsection we will apply *kernelization* to the data from the source and target domains and show that (7) can be solved in a kernel space by learning a nonlinear transformation metric with its size independent of the feature dimension.

### 2.3. Nonlinear Feature Transformation

Note that the size of the linear transformation metric $\mathbf{H}$ is proportional to the feature dimension, and thus it is computationally infeasible for data with a very high dimension. In this section, we will show that by applying kernelization, the transformation metric is independent of the feature dimension and grows *only* with the number of training data.

As any arbitrary feature mapping function $\phi$ can be used to derive a corresponding kernel space for the source and target data, we can replace their linear inner products with some kernel function $k$. Let us denote $\Phi_s = [\phi(\mathbf{x}_1^s), \ldots, \phi(\mathbf{x}_{n_s}^s)]$ and $\Phi_t = [\phi(\mathbf{x}_1^t), \ldots, \phi(\mathbf{x}_{n_t}^t)]$ as the matrices of the source and target training data after mapping them into a nonlinear feature space by using $\phi$, respectively. We also define $\mathbf{K}_s = \Phi_s' \Phi_s$ and $\mathbf{K}_t = \Phi_t' \Phi_t$ as the kernel matrices of the training data from the source and target domains, respectively. Moreover, we denote the corresponding projection matrices for the source and target data respectively as $\mathbf{P}_\phi$ and $\mathbf{Q}_\phi$.

**Theorem 1.** *Assume $\mathbf{K}_s$ and $\mathbf{K}_t$ be positive definite. There exist two matrices $\tilde{\mathbf{P}} \in \mathbb{R}^{d_c \times n_s}$ and $\tilde{\mathbf{Q}} \in \mathbb{R}^{d_c \times n_t}$ such that any feasible solution $\mathbf{P}_\phi$ and $\mathbf{Q}_\phi$ to the kernelized version of (2) can be written in the form of $\mathbf{P}_\phi = \tilde{\mathbf{P}} \mathbf{K}_s^{-1/2} \Phi_s'$ and $\mathbf{Q}_\phi = \tilde{\mathbf{Q}} \mathbf{K}_t^{-1/2} \Phi_t'$, respectively.*

*Proof.* The proof can be analogously derived as for Lemma 3.1 in (Kulis et al., 2011). □

With Theorem 1, it is easy to verify that $\|\tilde{\mathbf{P}}\|_F^2 = \|\mathbf{P}_\phi\|_F^2 \leq \lambda_p$ and $\|\tilde{\mathbf{Q}}\|_F^2 = \|\mathbf{Q}_\phi\|_F^2 \leq \lambda_q$. Here, we apply the same trick as in Section 2.2 to avoid determining $d_c$ for the latent common subspace. That is, we define the *nonlinear transformation metric* $\tilde{\mathbf{H}} = [\tilde{\mathbf{P}}, \tilde{\mathbf{Q}}]'[\tilde{\mathbf{P}}, \tilde{\mathbf{Q}}] \in \mathbb{R}^{(n_s+n_t) \times (n_s+n_t)}$, and thus the size of

---

**Algorithm 1** Heterogeneous Feature Augmentation

**Input:** Labeled source samples $\{(\mathbf{x}_i^s, y_i^s)|_{i=1}^{n_s}\}$ and labeled target samples $\{(\mathbf{x}_i^t, y_i^t)|_{i=1}^{n_t}\}$
**Initialization:** $\tau \leftarrow 1$, $\tilde{\mathbf{H}}_{[\tau]} \leftarrow \frac{\lambda}{n_s+n_t}\mathbf{I}_{n_s+n_t}$
With $\tilde{\mathbf{H}}_{[\tau]}$, solve for $\boldsymbol{\alpha}_{[\tau]}$ in the inner optimization problem of (8) by using SVM;
**while** $\tau < T_{\max}$ **do**
  Update $\tilde{\mathbf{H}}_{[\tau+1]}$ by using the projected gradient descent method with (10);
  With $\tilde{\mathbf{H}}_{[\tau+1]}$, solve for $\boldsymbol{\alpha}_{[\tau+1]}$ in the inner optimization problem of (8) by using SVM;
  **if** *the objective value of (8) converges* **then**
  | break;
  **end**
  $\tau \leftarrow \tau + 1$;
**end**
**Output:** $\tilde{\mathbf{H}}_{[\tau]}$ and $\boldsymbol{\alpha}_{[\tau]}$

---

$\tilde{\mathbf{H}}$ is independent of the feature dimension. We also have $\tilde{\mathbf{H}} \succeq 0$ and $\text{trace}(\tilde{\mathbf{H}}) \leq \lambda_p + \lambda_q = \lambda$.

Therefore, the formulation of our proposed HFA method after applying kernelization becomes:

$$\min_{\tilde{\mathbf{H}} \succeq 0} \max_{\boldsymbol{\alpha}} \quad \mathbf{1}'_{n_s+n_t} \boldsymbol{\alpha} - \frac{1}{2}(\boldsymbol{\alpha} \circ \mathbf{y})' \mathbf{K}_{\tilde{\mathbf{H}}}(\boldsymbol{\alpha} \circ \mathbf{y}), \quad (8)$$

s.t. $\mathbf{y}'\boldsymbol{\alpha} = 0$, $\mathbf{0}_{n_s+n_t} \leq \boldsymbol{\alpha} \leq C\mathbf{1}_{n_s+n_t}$, $\text{trace}(\tilde{\mathbf{H}}) \leq \lambda$,

where $\mathbf{K}_{\tilde{\mathbf{H}}} = \begin{bmatrix} \mathbf{K}_s + \tilde{\mathbf{L}}_s' \tilde{\mathbf{H}} \tilde{\mathbf{L}}_s & \tilde{\mathbf{L}}_s' \tilde{\mathbf{H}} \tilde{\mathbf{L}}_t \\ \tilde{\mathbf{L}}_t' \tilde{\mathbf{H}} \tilde{\mathbf{L}}_s & \mathbf{K}_t + \tilde{\mathbf{L}}_t' \tilde{\mathbf{H}} \tilde{\mathbf{L}}_t \end{bmatrix}$, $\tilde{\mathbf{L}}_s = \begin{bmatrix} \mathbf{I}_{n_s} \\ \mathbf{O}_{n_t \times n_s} \end{bmatrix} \mathbf{K}_s^{1/2}$ and $\tilde{\mathbf{L}}_t = \begin{bmatrix} \mathbf{O}_{n_s \times n_t} \\ \mathbf{I}_{n_t} \end{bmatrix} \mathbf{K}_t^{1/2}$. For a given $\boldsymbol{\alpha}$, we also arrive at an SDP problem as follows by defining $\boldsymbol{\beta} = \boldsymbol{\alpha} \circ \mathbf{y}$:

$$\min_{\tilde{\mathbf{H}} \succeq 0} -\frac{1}{2} \boldsymbol{\beta}' \mathbf{K}_{\tilde{\mathbf{H}}} \boldsymbol{\beta}, \quad \text{s.t. } \text{trace}(\tilde{\mathbf{H}}) \leq \lambda. \quad (9)$$

### 2.4. Detailed Solution

For our proposed HFA method, we develop an alternating optimization algorithm by iteratively updating $\boldsymbol{\alpha}$ and $\tilde{\mathbf{H}}$ to effectively solve the problem in (8). Specifically, when updating $\boldsymbol{\alpha}$ at the $\tau$-th iteration, we fix $\tilde{\mathbf{H}}_{[\tau]}$ and solve for $\boldsymbol{\alpha}_{[\tau]}$ in (8) by using the standard SVM with the kernel matrix $\mathbf{K}_{\tilde{\mathbf{H}}_{[\tau]}}$. While updating $\tilde{\mathbf{H}}$, we fix $\boldsymbol{\alpha}_{[\tau]}$ and solve for $\tilde{\mathbf{H}}_{[\tau+1]}$ via SDP optimization in (9). The optimization procedure will be terminated when the value of the objective function in (8) converges.

In order to efficiently solve the SDP problem in (9), we also develop a simple projected gradient descent method to update $\tilde{\mathbf{H}}$. Let us define $\boldsymbol{\beta}_s = [\beta_1, \ldots, \beta_{n_s}]'$ and $\boldsymbol{\beta}_t = [\beta_{n_s+1}, \ldots, \beta_{n_s+n_t}]'$. Denoting $G(\tilde{\mathbf{H}})$ as the objective function of (9), we first obtain the derivative of $G(\tilde{\mathbf{H}})$ with respect to $\tilde{\mathbf{H}}$ as follows:

$$\frac{\partial G}{\partial \tilde{\mathbf{H}}} = -\frac{1}{2}(\tilde{\mathbf{L}}_s\boldsymbol{\beta}_s + \tilde{\mathbf{L}}_t\boldsymbol{\beta}_t)(\tilde{\mathbf{L}}_s\boldsymbol{\beta}_s + \tilde{\mathbf{L}}_t\boldsymbol{\beta}_t)'.$$

Then at the $\tau$-th iteration, $\tilde{\mathbf{H}}$ will be updated by using the following equation:

$$\tilde{\mathbf{H}}_{[\tau+1]} = \tilde{\mathbf{H}}_{[\tau]} - \eta_{[\tau]} \frac{\partial G}{\partial \tilde{\mathbf{H}}}\bigg|_{\tilde{\mathbf{H}}=\tilde{\mathbf{H}}_{[\tau]}}, \quad (10)$$

where $\eta_{[\tau]}$ is the step size at the $\tau$-th iteration, which can be found by using the standard line search method (Boyd & Vandenberghe, 2004).

We summarize the proposed alternating optimization algorithm for HFA in Algorithm 1. After obtaining the optimal solution $\boldsymbol{\alpha}$ and $\tilde{\mathbf{H}}$ to (8), we can predict any test data point $\mathbf{x}$ from the target domain by using the following target decision function:

$$\begin{aligned}f(\mathbf{x}) &= \mathbf{w}'\phi(\mathbf{x}) + b \\ &= \left(\left(\boldsymbol{\beta}_s'\tilde{\mathbf{L}}_s' + \boldsymbol{\beta}_t'\tilde{\mathbf{L}}_t'\right)\tilde{\mathbf{H}}\begin{bmatrix}\mathbf{O}_{n_s \times n_t} \\ \mathbf{I}_{n_t}\end{bmatrix} + \boldsymbol{\beta}_t'\right)\mathbf{k}_t + b, (11)\end{aligned}$$

where $\mathbf{k}_t = [k(\mathbf{x}_1^t, \mathbf{x}), \ldots, k(\mathbf{x}_{n_t}^t, \mathbf{x})]'$ and $k(\mathbf{x}_i, \mathbf{x}_j) = \phi(\mathbf{x}_i)'\phi(\mathbf{x}_j)$ is a predefined kernel function for two data samples $\mathbf{x}_i$ and $\mathbf{x}_j$ with the same feature dimension.

## 3. Related Work

The pioneer works (Dai et al., 2009; Prettenhofer & Stein, 2010; Wei & Pal, 2010; Yang et al., 2009; Zhu et al., 2011) are limited to some specific HDA tasks, because they required additional information to transfer the source knowledge to the target domain.

To handle more general HDA tasks, other methods have been proposed to explicitly discover a common subspace (Shi et al., 2010; Wang & Mahadevan, 2011). Shi et al. (2010) proposed to learn feature mapping matrices without using the valuable data label information. While Wang et al. (2011) used the class labels of data, they assumed the data should have a manifold structure. Such manifold assumption may not exist in real-world applications.

Recently, Kulis et al. (2011) proposed a nonlinear metric learning method to learn an asymmetric feature transformation for the source and target data with high dimensions. And the learned transformation metric is universal for all classes. However, when there exist many classes, a universal metric may not be sufficiently good for feature transformation between data from all classes. In contrast, our method incorporates the proposed augmented features into SVM to learn an individual model for each class.

Table 1. Summarization of the object dataset with 31 categories.

|  | Domain | # dim | # total imgs | # training imgs per category |
|---|---|---|---|---|
| Source | amazon | 800 | 2,813 | 20 |
|  | webcam | 800 | 795 | 8 |
| Target | dslr | 600 | 498 | 3 |

## 4. Experiments

In this section, we evaluate our proposed HFA method for object recognition and multilingual text categorization. We focus on the heterogeneous domain adaptation problem where there exist only one source domain and one target domain in which only a limited number of labeled target training samples are available. Moreover, we assume that the test data from the target domain are unseen during the training phase.

### 4.1. Setup

**Object recognition:** We employ a recently released dataset[1] used in (Saenko et al., 2010; Kulis et al., 2011) for this task. This dataset contains a total of 4106 images with 31 categories from three sources: amazon (web images downloaded from an online merchant), dslr (high-resolution images taken from a digital DLR camera) and webcam (low-resolution images taken from a web camera). We follow the same protocols in the previous work (Kulis et al., 2011). Specifically, SURF features (Bay et al., 2006) are extracted for all the images. The images from amazon and webcam are clustered into 800 visual words by using k-means. After vector quantization, each image is represented as a 800 dimensional histogram feature. Similarly, we represented each image from dslr as a 600-dimensional histogram feature.

In the experiments, dslr is used as the target domain, while amazon and webcam are considered as two individual source domains. We strictly follow the setting in (Saenko et al., 2010; Kulis et al., 2011) and randomly select 20 (resp., 8) training images per category for the source domain amazon (resp., webcam). For the target domain dslr, 3 training images are randomly selected from each category, and the remaining dslr images are used for testing. See Table 1 for a summarization of this dataset.

**Text categorization:** We use the Reuters multilingual dataset[2] (Amini et al., 2009), which is collected by sampling parts of the Reuters RCV1 and RCV2 collections. It contains about 11K newswire articles from 6 classes in 5 languages (i.e., English, French, German, Italian and Spanish). While each document was also translated into the other four languages in this dataset, we do not use the translated documents in this work. All documents are represented as a bag of words and the TF-IDF are extracted.

---

[1] http://www.icsi.berkeley.edu/~saenko/projects.html

[2] http://multilingreuters.iit.nrc.ca/ReutersMultiLingualMultiView.htm

Table 2. Summarization of the Reuters multilingual dataset with 6 classes.

|  | Domain | # dim after PCA | # total docs | # training docs per class |
|---|---|---|---|---|
| Source | English | 1,131 | 18,758 | 100 |
|  | French | 1,230 | 26,648 | 100 |
|  | German | 1,417 | 29,953 | 100 |
|  | Italian | 1,041 | 24,039 | 100 |
| Target | Spanish | 807 | 11,547 | 5/7/10/15/20 |

We take Spanish as the target domain in the experiments and other four languages as individual source domains. For each class, we randomly sample 100 training documents from the source domain and $m$ training documents from the target domain, where $m = 5, 7, 10, 15$ and $20$. And the remaining documents in the target domain are used as the test data. Note that the method (Wang & Mahadevan, 2011) cannot handle the original high dimensional TF-IDF features. In order to compare our HFA method with theirs (Wang & Mahadevan, 2011), for documents written in each language, we perform PCA with 60% energy preserved on the TF-IDF features. We summarize this dataset in Table 2.

**Baselines:** As the source and target data have different dimensions, they cannot be directly combined to train any classifiers for the target domain. Considering the number of training samples is much lower than the feature dimension, we compare our HFA method by applying kernelization with a number of baseline algorithms listed below:

- **SVM_T:** It utilizes the labeled samples only from the target domain to train a standard SVM classifier for each category/class. Note it is not reported in (Kulis et al., 2011).

- **KCCA (Shawe-Taylor & Cristianini, 2004):** It learns a common feature subspace by maximizing the correlation between the source and target training data without using any label information. The data from both domains are projected into the common subspace. Note that KCCA was originally proposed for multi-view learning. Following (Kulis et al., 2011), we also report its results in this work.

- **HeMap (Shi et al., 2010):** It finds the projection matrices for a common feature subspace as well as learns the optimal projected data from both domains. But the label information of training data from both domains is not used.

- **DAMA (Wang & Mahadevan, 2011):** It learns a common feature subspace by utilizing the class labels of the source and target training data for manifold alignment.

- **ARC-t (Kulis et al., 2011):** It uses the labeled training data from both domains to learn an asymmetric transformation metric between the different feature spaces.

For KCCA, HeMap and DAMA, after learning the projection matrices, we apply SVM to train their final classifiers by using the projected training data from both domains for a given category/class. For ARC-t, we construct the kernel matrix based on the learned asymmetric transformation metric, and then SVM is also applied to train its final classifier. For all methods, we set the regularization parameter $C = 1$ in SVM and use the RBF kernel for fair comparison. As we only have a very limited number of labeled training samples in the target domain, the cross-validation technique cannot be effectively employed to determine the optimal parameters. Instead, for our HFA method, we empirically fix the parameter $\lambda$ as 100 for the object dataset and 1 for the Reuters multilingual dataset. For other methods, we validate all their parameters chosen from $\{0.01, 0.1, 1, 10, 100\}$ based on the test data and report their best results.

**Evaluation metric:** Following (Kulis et al., 2011), for each method we measure the classification accuracy over all categories/classes on both datasets. We randomly sample the training data for ten times and report the mean classification accuracies of all methods over the ten rounds of experiments.

### 4.2. Classification Results

**Object recognition:** We report the mean and standard deviations of classification accuracies for all methods on the object dataset (Saenko et al., 2010) in Table 3. From the results, SVM_T outperforms KCCA and HeMap by using only 3 labeled training samples from the target domain. The explanation is that KCCA and HeMap do not utilize the label information of the target training data to learn the feature mapping matrices. As a result, the learned common subspace is not sufficient to preserve a similar data structure as in the original feature spaces of the source and target data, which results in poor classification performances. DAMA performs only slightly better that SVM_T, possibly due to the lack of

Table 3. Means and standard deviations of classification accuracies (%) of all methods on the object dataset by using 3 labeled training samples per class from the target domain dslr. Results in boldface are significantly better than the others, judged by the t-test with a significance level at 0.05. For KCCA and ARC-t, the numbers in the parentheses are the results reported in (Kulis et al., 2011).

| Source Domain | SVM_T | KCCA | HeMap | DAMA | ARC-t | HFA |
|---|---|---|---|---|---|---|
| amazon | $52.9 \pm 3.1$ | $46.3 \pm 2.7$ (51.0) | $42.8 \pm 2.4$ | $53.3 \pm 2.3$ | $53.1 \pm 2.4$ (53.2) | $\mathbf{55.4 \pm 2.8}$ |
| webcam | | $46.7 \pm 2.8$ | $42.2 \pm 2.6$ | $53.2 \pm 3.2$ | $53.0 \pm 3.2$ | $\mathbf{54.3 \pm 3.7}$ |

Table 4. Means and standard deviations of classification accuracies (%) of all methods on the Reuters multilingual dataset by using 10 labeled training samples per class from the target domain Spanish. Results in boldface are significantly better than the others, judged by the t-test with a significance level at 0.05.

| Source Domain | SVM_T | KCCA | HeMap | DAMA | ARC-t | HFA |
|---|---|---|---|---|---|---|
| English | $72.6 \pm 2.3$ | $71.4 \pm 3.2$ | $65.7 \pm 3.1$ | $72.4 \pm 2.4$ | $72.9 \pm 2.0$ | $\mathbf{75.3 \pm 1.7}$ |
| French | | $72.8 \pm 2.8$ | $64.2 \pm 4.2$ | $72.8 \pm 2.0$ | $73.5 \pm 1.8$ | $\mathbf{75.7 \pm 1.6}$ |
| German | | $73.8 \pm 2.2$ | $64.6 \pm 3.6$ | $72.9 \pm 2.3$ | $74.7 \pm 1.6$ | $\mathbf{76.1 \pm 1.5}$ |
| Italian | | $73.8 \pm 2.1$ | $65.8 \pm 2.3$ | $73.3 \pm 2.1$ | $74.0 \pm 2.0$ | $\mathbf{75.8 \pm 1.8}$ |

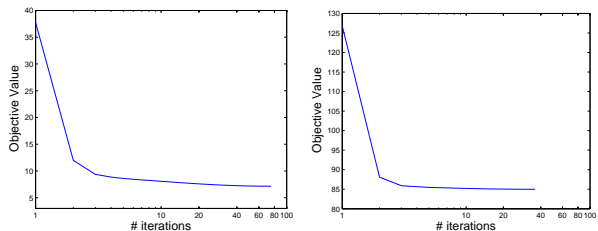

(a) "back_pack" on the object dataset. (b) "C15" on the Reuters multilingual dataset.

Figure 1. Illustrations of the convergence of Algorithm 1 for our HFA method on the two datasets.

the strong manifold structure on this dataset. Both results of ARC-t implemented by ourselves and reported in (Kulis et al., 2011) are only comparable with those of SVM_T, which shows that ARC-t is less effective for HDA on this dataset. Our HFA method outperforms the other methods under both settings, which clearly demonstrate the effectiveness of our proposed method for HDA by learning with augmented features.

**Text categorization:** Table 4 shows the mean and standard deviations of classification accuracies for all methods on the Reuters multilingual dataset (Amini et al., 2009) by using $m = 20$ labeled training samples per class from the target domain. We have a similar observation as on the object dataset that SVM_T still outperforms HeMap in terms of classification accuracy. It is interesting to observe that KCCA is generally better than SVM_T, which shows that it can learn a good common feature subspace on this dataset. Moreover, by using the label information, both DAMA and ARC-t perform better than SVM_T under almost all the settings. Our proposed HFA method still achieves significantly better performances than others on this dataset, when judged by the t-test with a significance level at 0.05.

We also plot the classification results of SVM_T, KCCA, DAMA, ARC-t and our HFA method with respect to the number of target training samples per class (i.e., $m = 5, 7, 10, 15$ and $20$) for each source domain in Figure 2. We do not report the results of HeMap, as they are much worse than the other methods. From the results, the performances of all methods increase when using a larger $m$. And the two HDA methods DAMA and ARC-t generally achieve better mean classification accuracies than SVM_T except for the setting using English as the source domain. Our HFA method generally outperforms all other methods according to mean classification accuracy.

### 4.3. Convergence Analysis

To analyze the convergence of the proposed Algorithm 1 for our HFA method, we take one setting from each of the datasets as the showcase. For the object dataset, we use the category "back_pack" and the source domain amazon; for the Reuters multilingual dataset, the class "C15" is used together with the source domain English. From the results, Algorithm 1 generally takes less than 80 (resp., 40) iterations before its convergence on the object dataset (resp., the Reuters multilingual dataset). We have similar observations for other categories/classes on the two datasets as well.

## 5. Conclusions and Future Work

We have proposed a new method called Heterogeneous Feature Augmentation (HFA) for heterogeneous domain adaptation. In HFA, we augment the heterogeneous features from the source and target domains by using two newly proposed feature mapping functions, respectively. With the augmented features, we propose to find the two projection matrices for the source and target data by using the standard SVM with the hinge loss in both linear and nonlinear cases. Moreover, a so-called transformation metric is introduced to simply our formulated optimization problem of HFA such that it can be effectively solved by our developed alternating optimization algorithm. Promising results of HFA have been achieved on two benchmark datasets for object recognition and text classification.

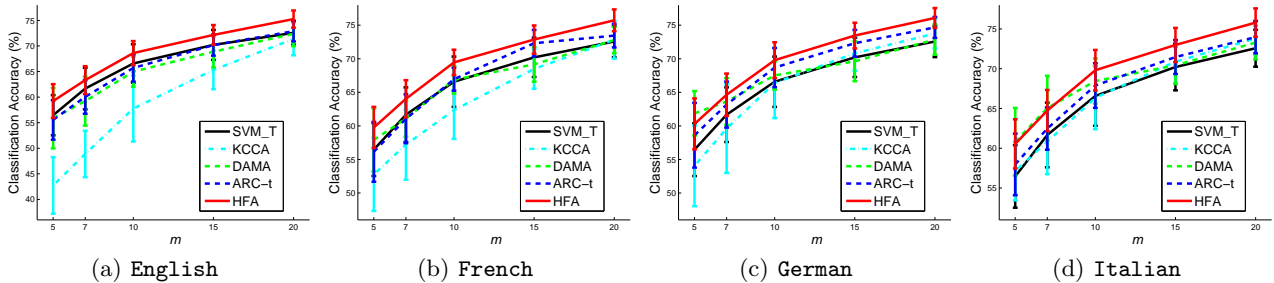

(a) English  (b) French  (c) German  (d) Italian

Figure 2. Classification accuracies of all methods with respect to different number of target training samples per class (i.e., $m = 5, 7, 10, 15$ and $20$) on the Reuters multilingual dataset. Spanish is considered as the target domain, and in each subfigure the results are obtained by using one language as the source domain.

## Acknowledgement

This work is supported by the Singapore National Research Foundation under its Interactive & Digital Media (IDM) Public Sector R&D Funding Initiative (administered by the IDM Programme Office) and AcRF Tier-1 Research Grant (RG15/08).

## References

Amini, M., Usunier, N., and Goutte, C. Learning from multiple partially observed views – an application to multilingual text categorization. In *NIPS*, 2009.

Bay, H., Tuytelaars, T., and Gool, L. Van. Surf: Speeded up robust features. In *ECCV*, 2006.

Blitzer, J., McDonald, R., and Pereira, F. Domain adaptation with structural correspondence learning. In *EMNLP*, 2006.

Blitzer, J., Dredze, M., and Pereira, F. Biographies, bollywood, boom-boxes and blenders: Domain adaptation for sentiment classification. In *ACL*, 2007.

Boyd, Stephen and Vandenberghe, Lieven. *Convex Optimization*. Cambridge University Press, 2004.

Dai, W., Chen, Y., Xue, G.-R., Yang, Q., and Yu, Y. Translated learning: Transfer learning across different feature spaces. In *NIPS*, 2009.

Daumé III, H. Frustratingly easy domain adaptation. In *ACL*, 2007.

Duan, L., Xu, D., Tsang, I. W., and Luo, J. Visual event recognition in videos by learning from web data. In *CVPR*, 2010.

Duan, L., Tsang, I. W., and Xu, D. Domain transfer multiple kernel learning. *T-PAMI*, 34(3):465–479, March 2012a.

Duan, L., Xu, D., and Tsang, I. W. Domain adaptation from multiple sources: A domain-dependent regularization approach. *T-NNLS*, 23(3):504–518, March 2012b.

Harel, M. and Mannor, S. Learning from multiple outlooks. In *ICML*, 2011.

Kulis, B., Saenko, K., and Darrell, T. What you saw is not what you get: Domain adaptation using asymmetric kernel transforms. In *CVPR*, 2011.

Pan, S. J., Ni, X., Sun, J.-T., Yang, Q., and Chen, Z. Cross-domain sentiment classification via spectral feature alignment. In *WWW*, 2010.

Prettenhofer, P. and Stein, B. Cross-language text classification using structural correspondence learning. In *ACL*, 2010.

Saenko, K., Kulis, B., Fritz, M., and Darrell, T. Adapting visual category models to new domains. In *ECCV*, 2010.

Shawe-Taylor, J. and Cristianini, N. *Kernel Methods for Pattern Analysis*. Cambridge University Press, 2004.

Shi, X., Liu, Q., Fan, W., Yu, P. S., and Zhu, R. Transfer learning on heterogenous feature spaces via spectral transformation. In *ICDM*, 2010.

Vandenberghe, L. and Boyd, S. Semidefinite programming. *SIAM Review*, 38(1):49–95, March 1996.

Wang, C. and Mahadevan, S. Heterogeneous domain adaptation using manifold alignment. In *IJCAI*, 2011.

Wei, B. and Pal, C. Cross-lingual adaptation: An experiment on sentiment classifications. In *ACL*, 2010.

Wu, P. and Dietterich, T. G. Improving svm accuracy by training on auxiliary data sources. In *ICML*, 2004.

Yang, Q., Chen, Y., Xue, G.-R., Dai, W., and Yu, Y. Heterogeneous transfer learning for image clustering via the social web. In *ACL/IJCNLP*, 2009.

Zhu, Y., Chen, Y., Lu, Z., Pan, S. J., Xue, G.-R., Yu, Y., and Yang, Q. Heterogeneous transfer learning for image classification. In *AAAI*, 2011.